# Heading Lock Maneuver Testing of Autonomous Underwater Vehicle : Sotong - ITB

**Muljowidodo K.*** and **Sapto Adi N.****

*Automation & Robotics Laboratory,*

*Mechanical Engineering Department - Bandung Institute of Technology**

*Center for Unmanned System Study*

*Bandung Institute of Technology***

*Jl. Ganesha 10, Bandung 40132, Indonesia*

Electronic mail: muljo@bdg.centrin.net.id

**Abstract**

In recent years, Autonomous Underwater Vehicle (AUV) research and development at Bandung Institute of Technology in Indonesia has achieved the testing stage in the field. This testing was still being classified as the early testing, since some of the preliminary tests were carried out in the scale of the laboratory. The paper would discuss the laboratory test and several tests that were done in the field. Discussions were stressed in the procedure and the aim that will be achieved, along with several early results. The testing was carried out in the lake with the area around 8300 Ha and the maximum depth of 50 meters. The location of the testing was chosen with consideration of minimizing the effect of the current and the wave, as well as the location that was not too far from the Laboratory. The type of testing that will be discussed in paper was Heading Lock Maneuver Testing. The vehicle was tested to move with a certain cruising speed, afterwards it was commanded by an arbitrarily selected heading direction. The response and the behavior of the vehicle were recorded as the data produced by the testing.

## 1 Introduction

Autonomous underwater vehicle –Sotong ITB was developed since 2004. Since that experienced various testing from the structure to functional in the scale of the laboratory. And in this year began to be carried out by the testing in the field. In this opportunity was expected to be able to be received by several results of the comparison towards the design originally, From the loading dynamically against the structure or system control from the vehicle. This was the configuration when being carried out by the testing:

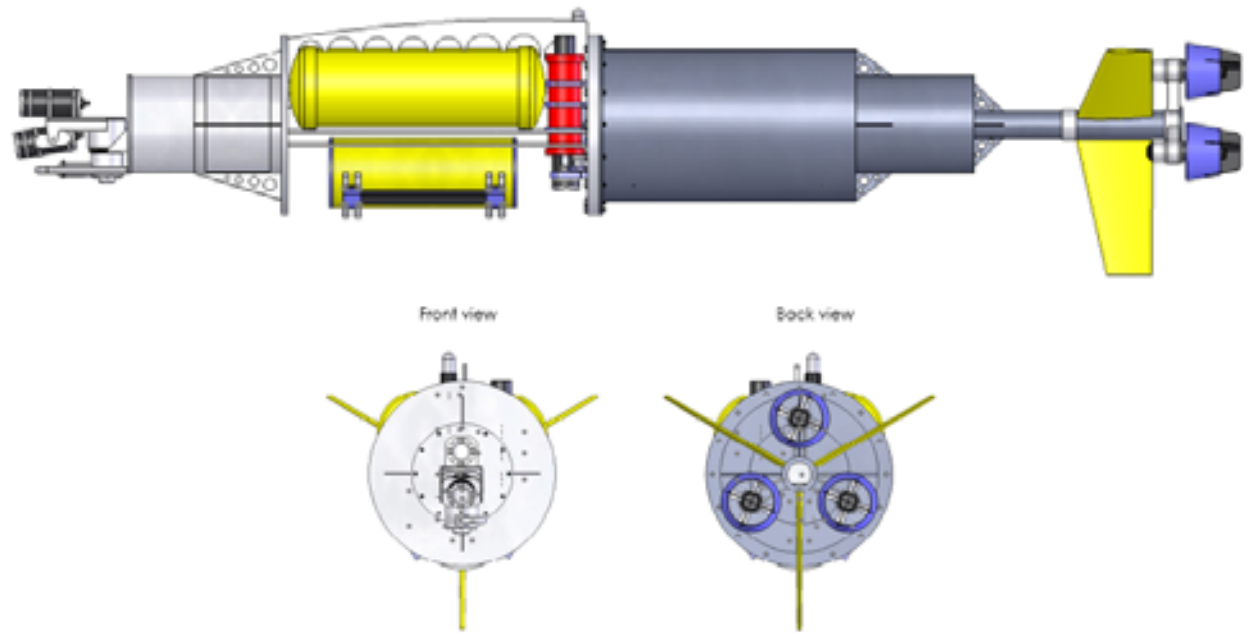

Figure1 Sotong Configurations on Field testing.

Three thruster with their respective maximum capacity 300 N (Bollard) ,and three stabilizer fin was the propulsion system for maneuver the vehicle. Classical PID Control was used in the testing this time.

This is the details from the configuration of the vehicle during the testing:

| Configuration | specifications |
|---|---|
| Length | 4400mm |
| Hull Diameter | 750mm |
| Displacement | 378 Kg |
| Weight on Air | 370 Kg |
| Navigation | USBL tracking system, Obstacle Avoidance sonar, Inertial Measurement Unit |
| Control | Classical PID |
| Propulsion | 3 Vector Thruster (@ 1 Hp, 300N ) |
| Payload | Side Scan Sonar |
| Power | 7.2 AH, 150VDC, Lead Acid |
| Emergency | Drop weight, Radio Beacon |

## 2 Testing Procedures

Before carrying out the testing in the field, several components were tested in the scale of the laboratory. One of them was Thruster.

The Thrust testing from the three thruster was aimed to receive the characteristics and their respective performance.

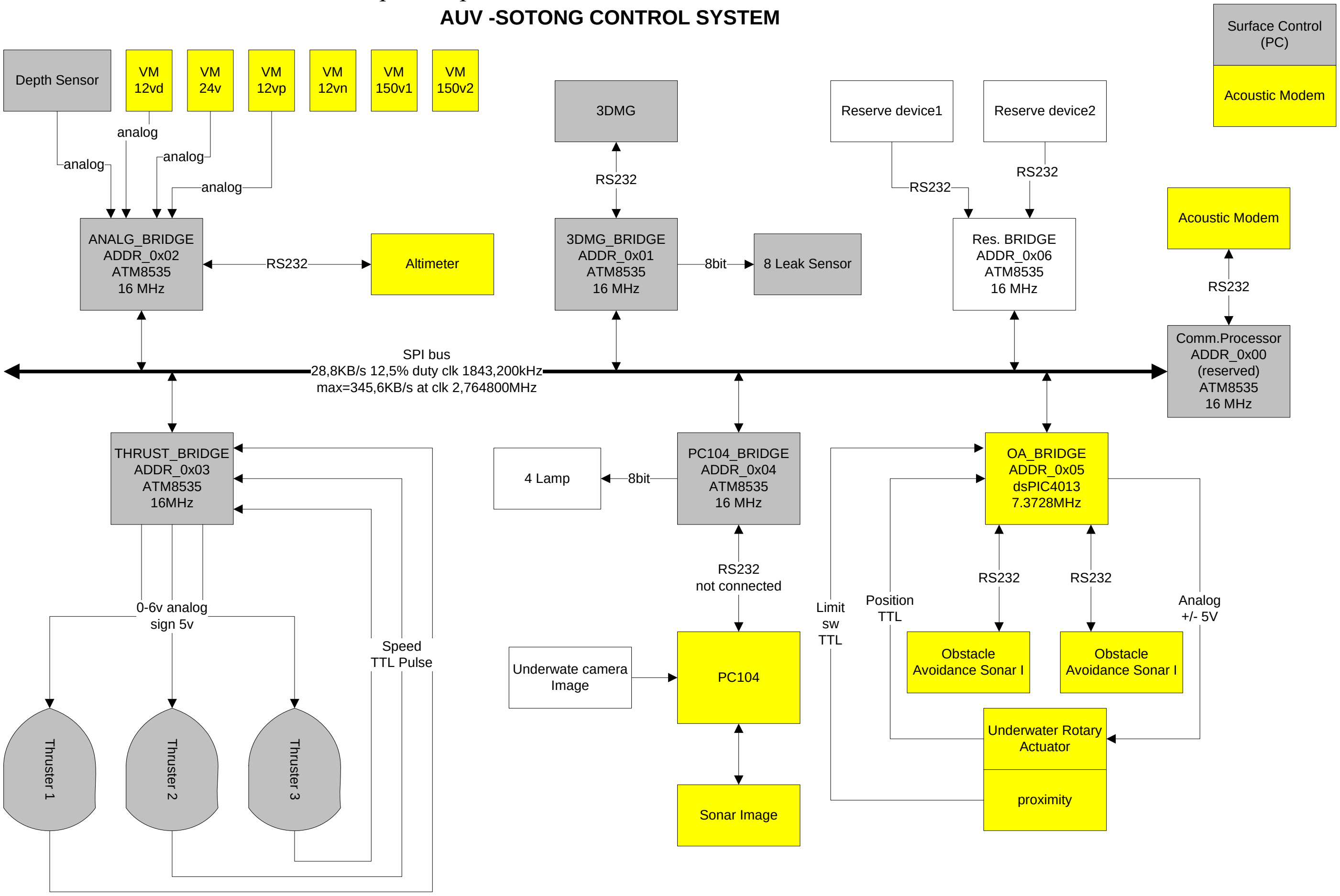

For the system control, this testing to know the controlling thruster Dead Zone.

The testing installation was shown in **figure 2**.

Results of the testing in the plot in the graph in **figure3.**

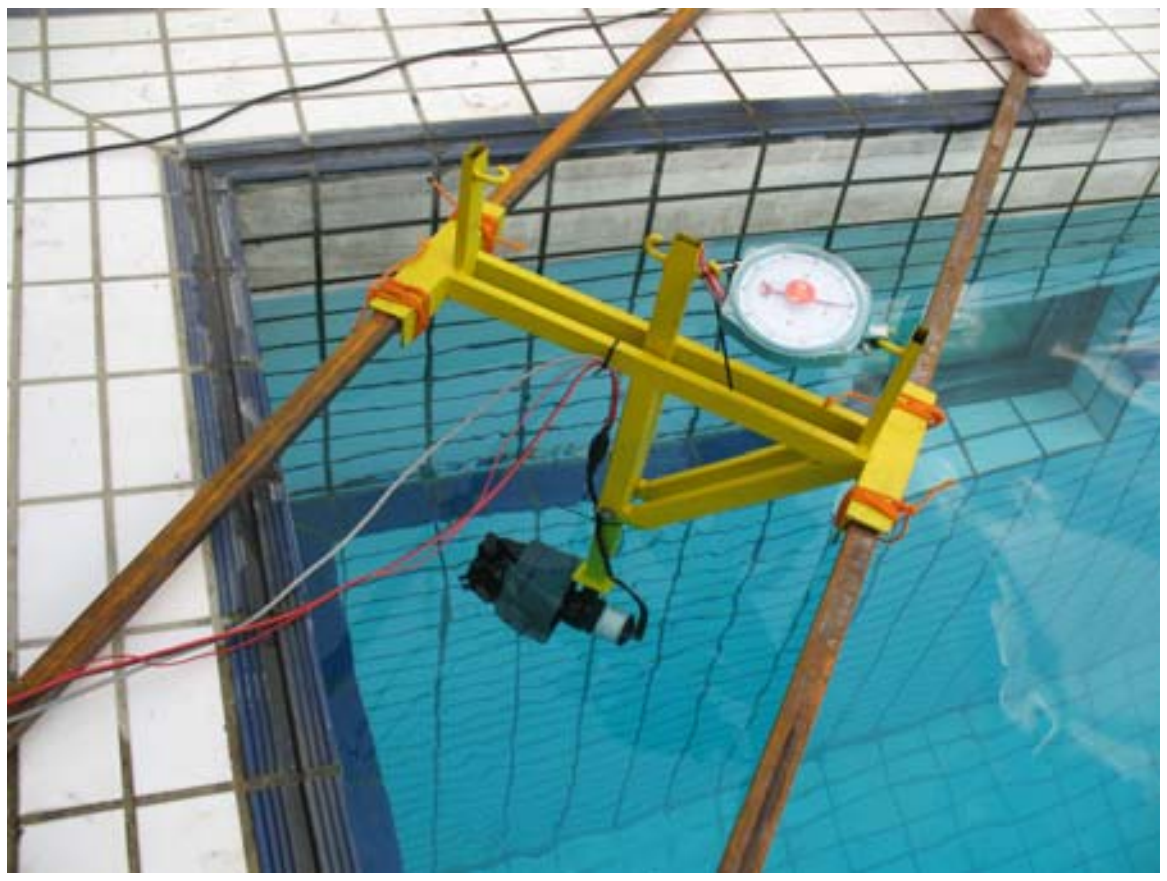

Figure 2. Thruster testing installation

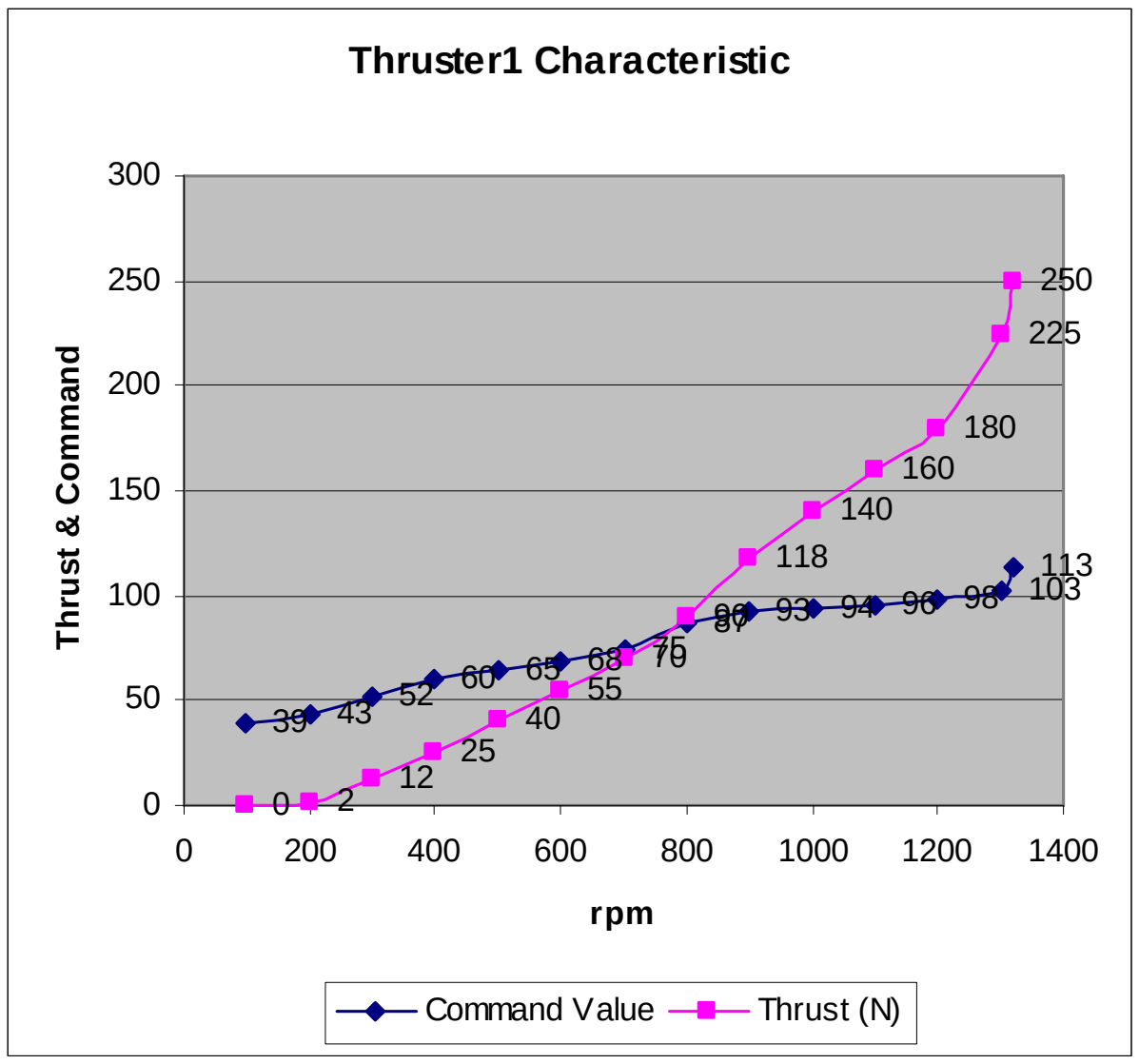

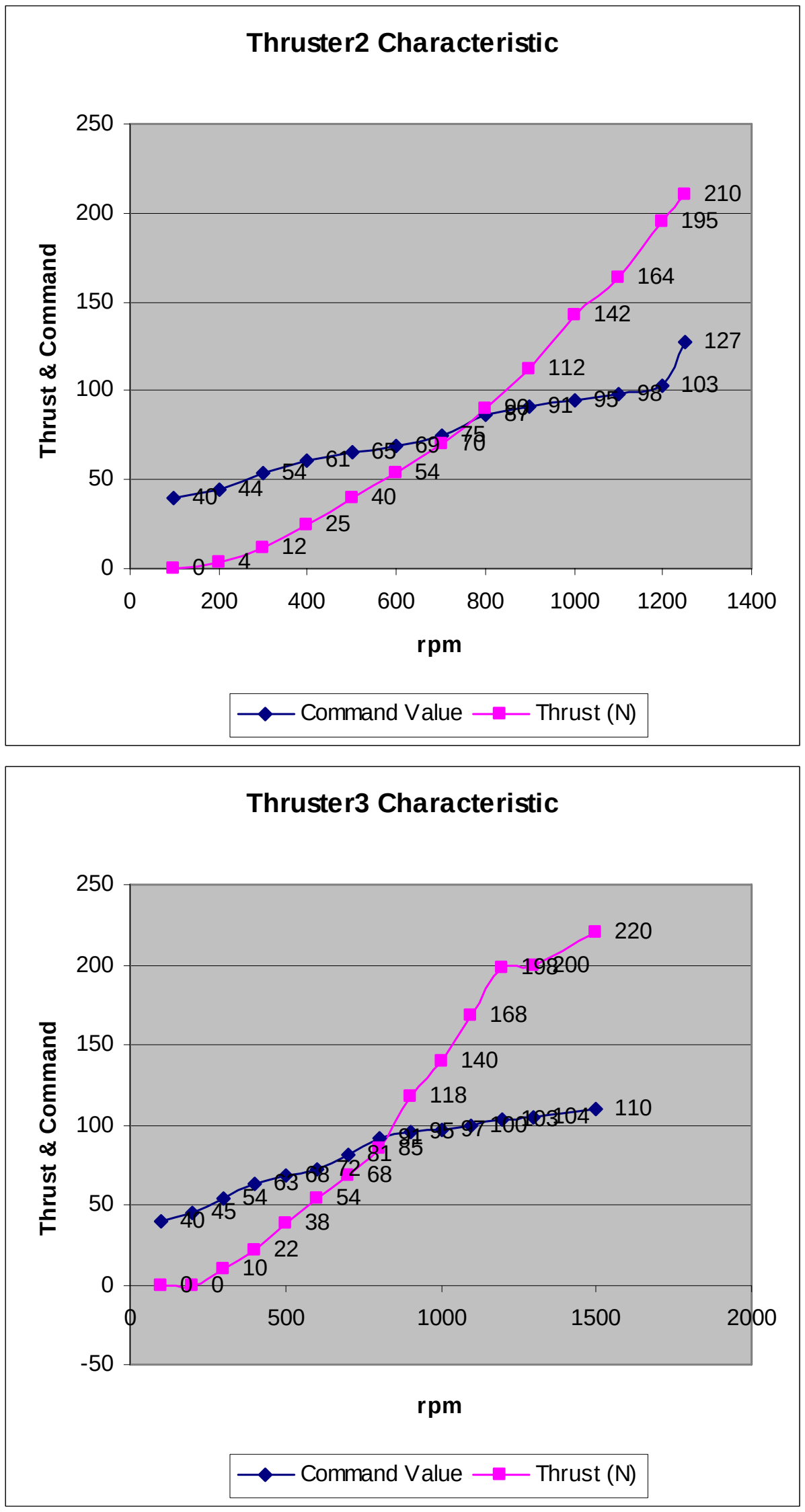

Figure3. Thruster Characterististic and Performance

The testing was carried out for 3 days with estimated if the weather that was not wanted happening so as did not enable was carried out by the testing. As in the case of rain or the noisiness of waters.
Before being carried out by the testing, before that was carried out by the ballast trimming so as to be obtained the pitching angle early around 0 deg.
This was meant to trace Outside Force that resulted in the change in the pitching angle.
The misalignment on fin stabilizer and asymmetrically from fairing will result in unbalance of hydrodynamic forces. In this testing, communication between the vehicle and surface control station in support vessel still used the cable (tethered).
Because bandwidth from underwater acoustic modem did not sufficient for the process of diagnosis mode that's needing the data current of two directions with 35 Hz Data update. Attitude (Roll, Pitch,& yaw), Depth, Altitude, Obstacle, Speed of each thruster (rpm), 8 position of leak sensor, & Voltage monitoring from power supply to All Devices must be recorded as long as the testing on going. With the existence of the cable, possibly will give the effect of the increase drag the vehicle.
Drag that was produced the cable with the diameter 8 mm and long around 10 meter, was not so significant if compared with thrust maks. of 900 N. Deployment this vehicle used the special design trailer so as all of his body could be unloaded directly in water without needing one crane or other adopted aids.

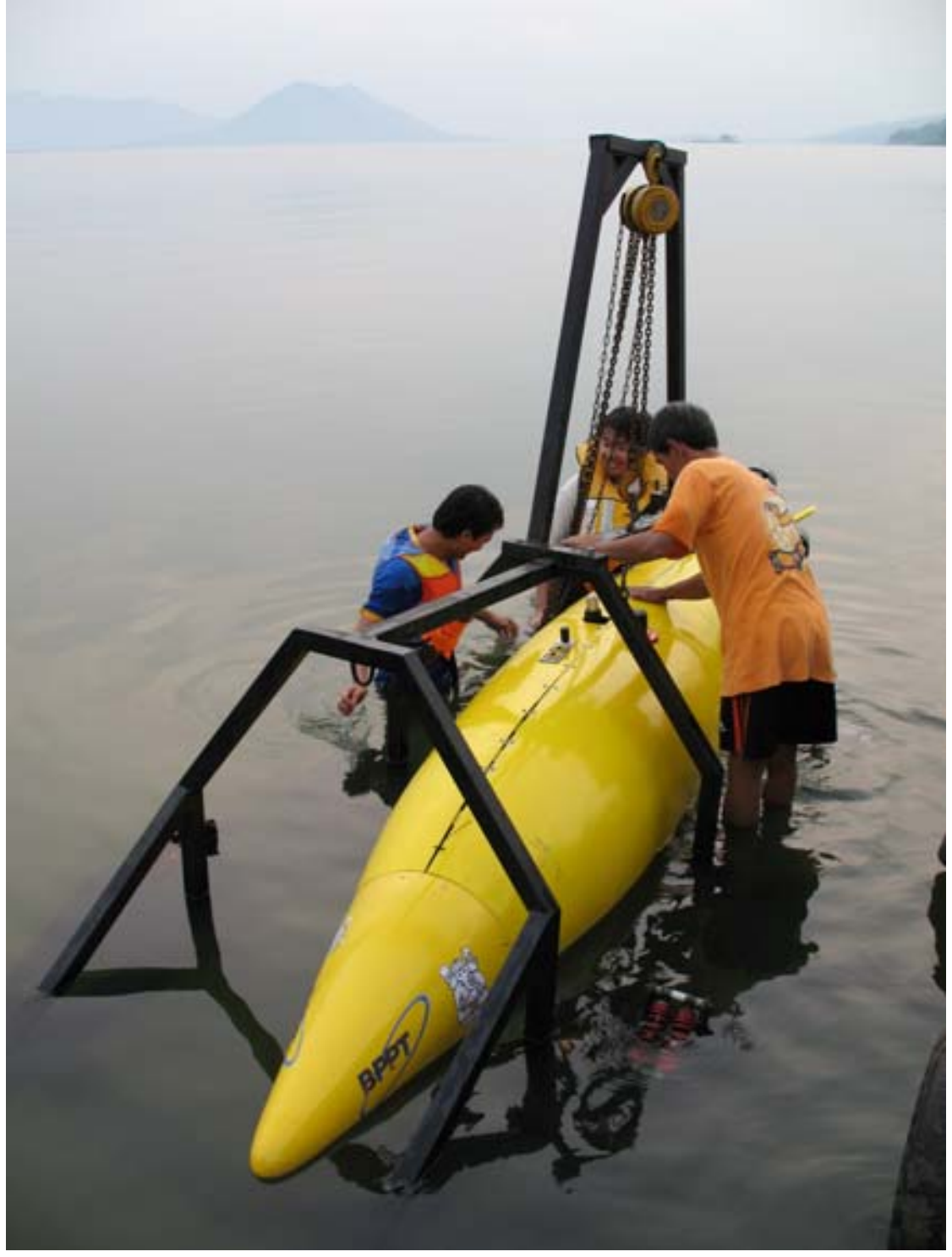

Figure 4 Sotong AUV deployment on Jatiluhur Bay

AUV – Sotong was controlled manually to maneuver certain, for example forward maneuver. Afterwards was given by the direction heading just and the predictions constant control was given randomly also.
The first trial: thrust early set 30 %, afterwards the pilot gave the movement forward maneuver. To the speed steady certain, control the manual was released and afterwards control automatic in activated with before gave the value of the constant control.
In the testing this time just was tried with the proportional constant from the PID whole the constant.**Fig. 5**

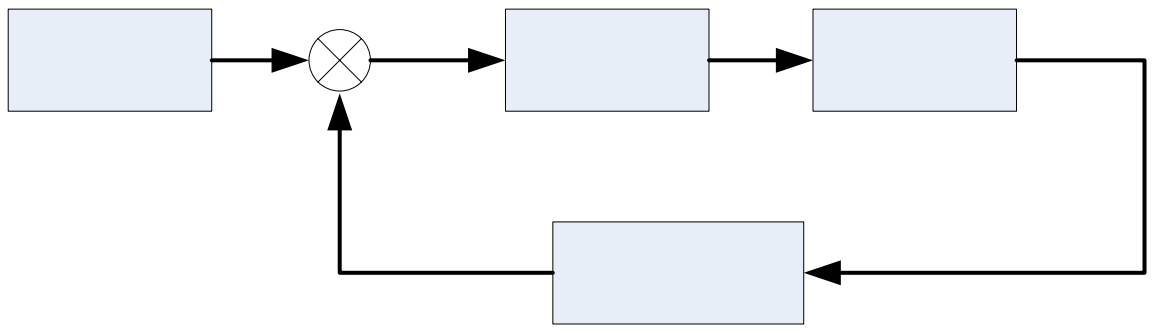

Figure 5. Control Schematic of Heading Lock maneuver

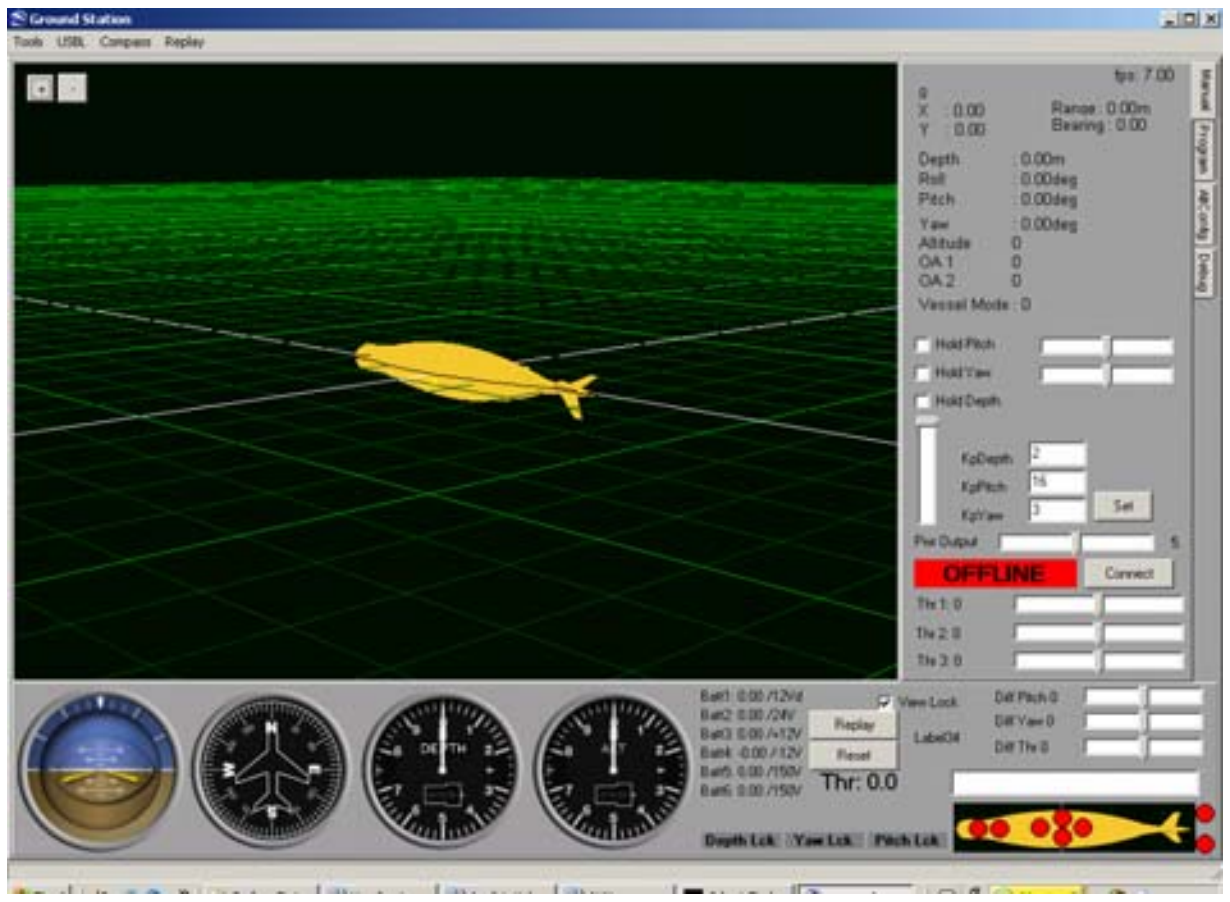

Figure 6. View from Control surface station computer

## 3 Testing Results

By being enough gave twice tuning the proportional constant, the vehicle could have responded headed heading that was determined. Despite still had the effect of indolence from the vehicle to do maneuver beforehand (Forward maneuver), step by step began to head to the side of heading actually. Along with this was the Heading Angle graph towards time when did maneuver (**figure 7**).

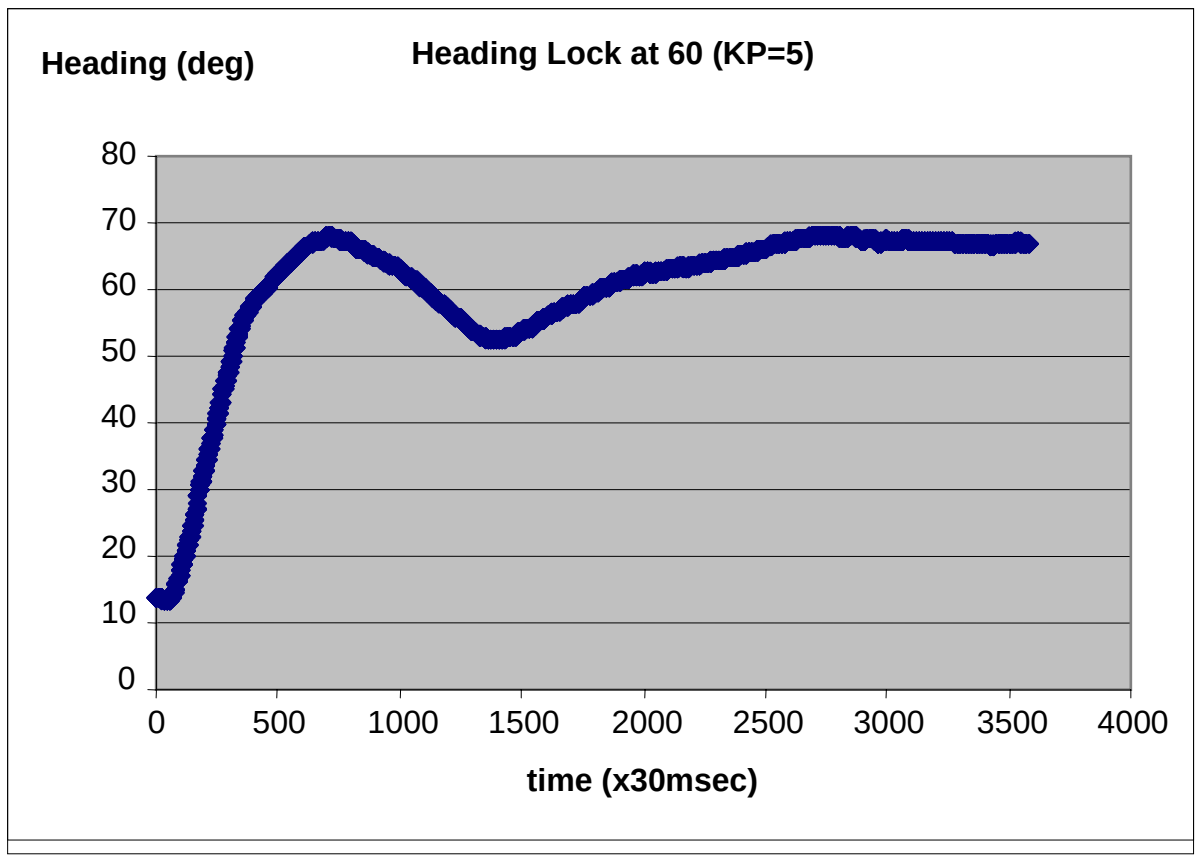

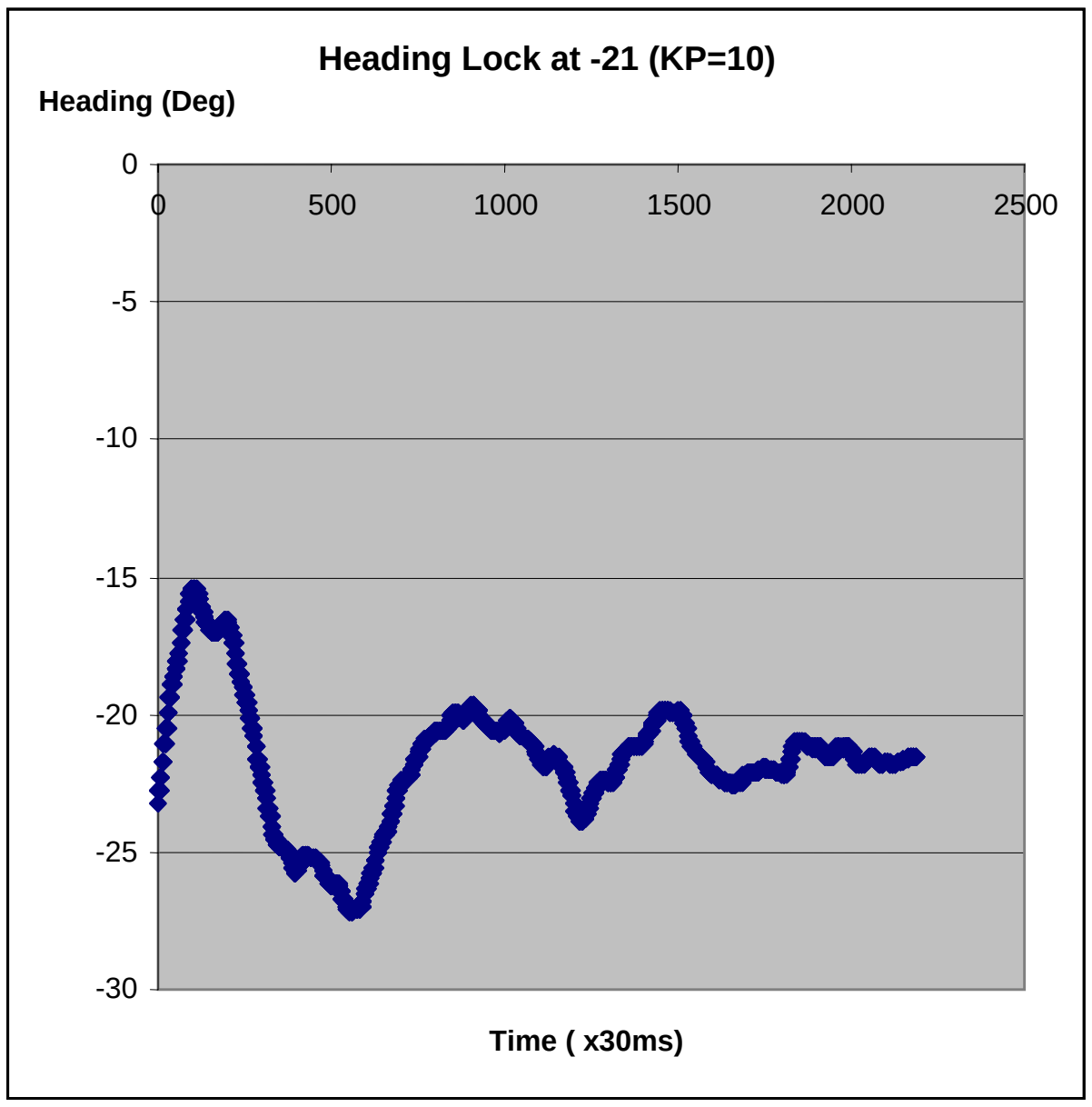

Figure 7. Heading Angle toward time graph

From two graph above, can be compared the performance of each constant. With KP=5, the Oscillation around destination heading angle could be avoided. But indolence phenomenon occurred to aim the position (Settling time). And steady state error recorded around 8 deg. That's looked different with the KP=10, although the oscillations still detected, but the time to reach the destination heading more shorter. If compared between both constant (KP) above with the same sampling error to destination heading, KP=10 more faster around 500 x 30 ms = 15000 ms= 15 sec.

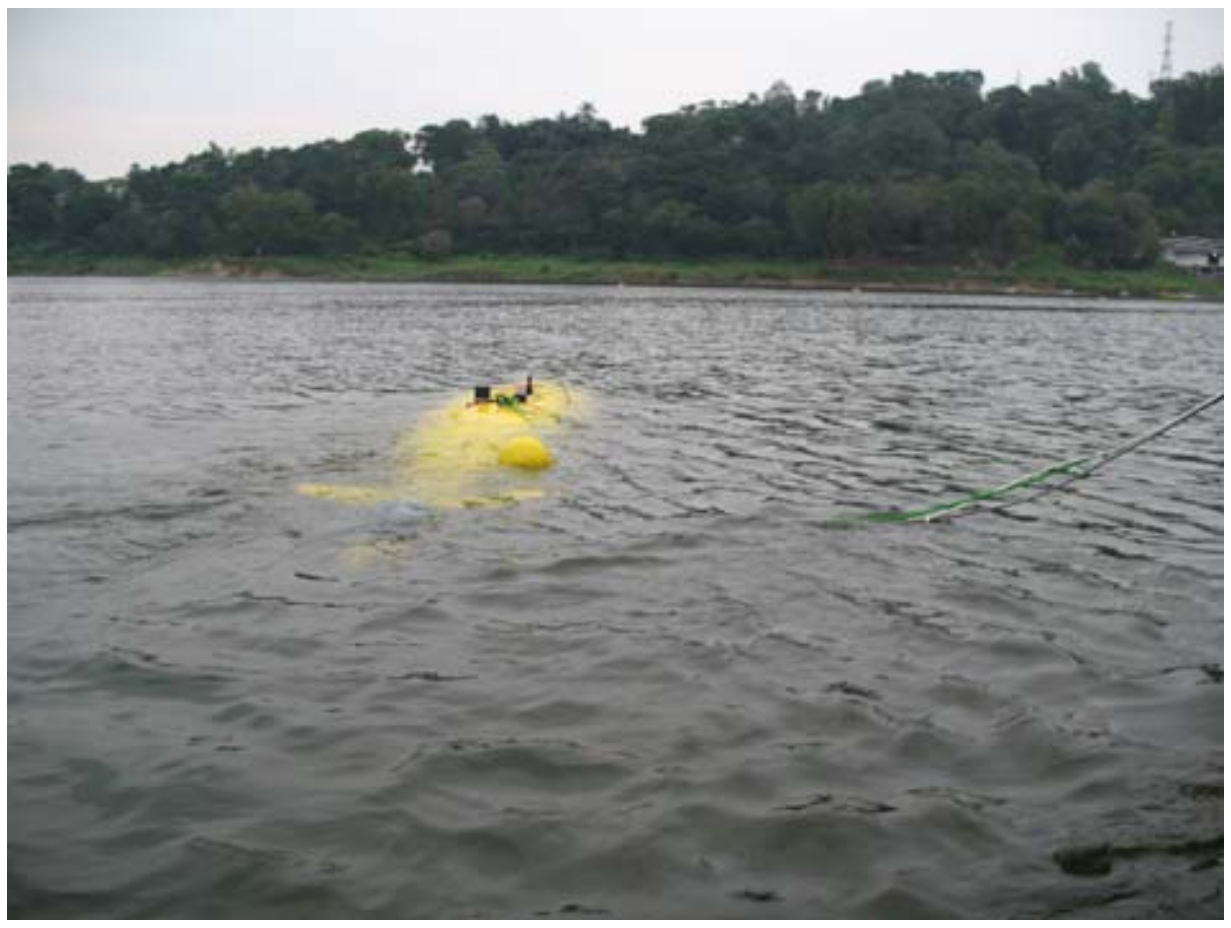

Figure 8. AUV-SOTONG on Maneuver

## 4 Technology Review of Underwater Vehicle Control

Due to the complex nature of the underwater vehicle dynamics, oceanic disturbance and uncertainty pertaining to changes in center of gravity and buoyancy,

AUVs demand control system that has a self-tuning ability. Numerous approaches to control strategy have been developed to address the need including sliding mode control, neural network, fuzzy control and adaptive control. Ref[7] reports comparative study of various control techniques for underwater vehicle used in the 1990s. The PID control was reported to be successful in the provision that the vehicle is operated under a constant speed thus justifying the linear approach. The comparison between $H_\infty$ controller and classical approach reveals that the two achieve similar performance making the former technique unattractive due to significantly more complex design procedure. The study concludes that the more obvious control candidates for initial investigation are classical controllers, fuzzy logic and sliding mode.

Classical PID control could be optimized with accurate movement system modeling method. The general underwater vehicle coordinate system can be shown on **figure 9** below [1].

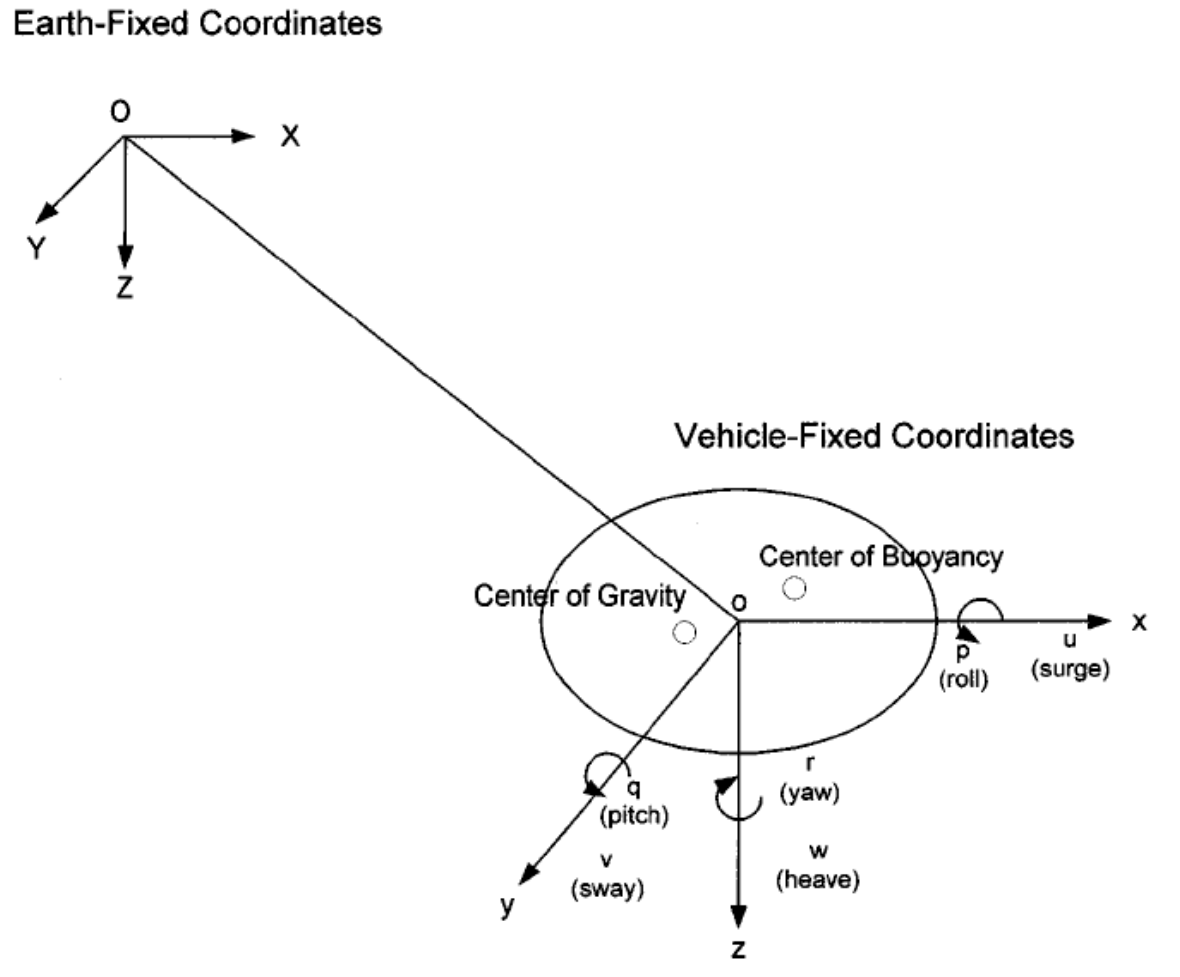

Figure 9. The Coordinate system for general underwater vehicle

Dynamics of underwater vehicles, including hydrodynamic parameter uncertainties, are highly nonlinear, coupled, and time varying. Several modeling and system identification techniques for underwater robotic vehicles have been proposed by researchers (Fossen, 1995; Goheen, 1995).

The effect of thruster dynamics on the vehicle also becomes significant, especially when the vehicle has slow and fine motion (Yoerger et al., 1990). Therefore, accurate modeling and verification by simulation are required steps in the design process (Lewis et al., 1984; Pappas et al.,1991). Integrated simulation (or HILS: hardware in the loop simulation) with actual parts of the vehicle and the environment is more desirable than completely numerical stand-alone simulation. Integrated simulation packages, including 3D graphics and virtual reality capabilities, are useful for developing advanced underwater robotic vehicles since actual field-testing is very expensive (Choi and Yuh, 1993; Brutzman et al., 1992; Kuroda et al., 1995).

The six degrees-of-freedom nonlinear equations of motion of the vehicle are defined with respect to two coordinate systems as shown in **Figure 9.** The vehicle coordinate system has six velocity components of motion (surge, sway, heave, roll, pitch, and yaw). The velocity vector in the vehicle coordinate system is expressed as $\dot{q} = [uvwpqr]^T$. The global coordinate system OXYZ is a fixed coordinate system. Translational and rotational movement in the global reference frame are represented by $x = [xyz\phi\theta\psi]^T$ that includes earth fixed positions and Euler angles. The equations of motion for underwater robots written as follows:

$$\dot{x} = J(x)\dot{q}$$

$$M\ddot{q} + C(\dot{q})\dot{q} + D(\dot{q})\dot{q} + G(x) = \tau + w$$

$$\tau = Bu$$

where ***J*(*x*)** is a 6x6 velocity transformation matrix that transforms velocities of the vehicle-fixed to the earth-fixed reference frame; ***M*** is a 6x6 inertia matrix as a sum of the rigid body inertia matrix $M_R$ and the hydrodynamic virtual inertia (added mass)$M_A$; is a 6x6 Coriolis and centripetal matrix including rigid body terms $C_R(\dot{q})$ and terms $C_A(\dot{q})$ due to added mass; $D(\dot{q})$ is a 6x 6 damping matrix including terms due to drag forces; ***G*(*x*)** is a 6x1 vector containing the restoring terms formed by the vehicle's buoyancy and gravitational terms; $\tau$ is a 6x1 vector including the control forces and moments; ***w*** is a 6x1 disturbance vector representing the environmental forces and moments(e.g. wave and current) acting on the vehicle; ***B*** is a control matrix of appropriate dimensions; and ***u*** is a vector whose components are thruster forces[1].

As the robot moves underwater, additional force and moment coefficients are added to account for the effective mass of the fluid that surrounds the robot and must be accelerated with the robot. These coefficients are referred to as added (virtual) mass and include added
moments of inertia and cross coupling terms such as
force coefficients due to linear and angular accelerations. The hydrodynamic added mass may be written with the SNAME (The Society of Naval Architects and Marine Engineers) convention such that for the hydrodynamic added mass force $Y_A$ along the *y*-axis due to a linear acceleration $\dot{u}$ in the *y*-direction is shown $y_A = -Y_{\dot{u}}.\dot{u}$ where $Y_{\dot{u}} = \partial Y / \partial \dot{u}$. Triantafyllou and Amzallag (1984) discussed how to calculate the various elements in$M_A$ for different geometrical bodies. In an ideal fluid, $M_A$ is strictly positive and symmetrical.

Based on the kinetic energy of the fluid, $E = \dot{q}^T M_A \dot{q} / 2$ ,the added mass forces and moments can be derived by Kirchhoff's equations (Kirchhoff, 1869; Sagatun,1992). Then, the added mass forces and moments can be seen as a sum of hydrodynamic inertia forces and moments $M_A\ddot{q}$ and hydrodynamic Coriolis and centripetal forces and moments $C_A(\dot{q})\dot{q}$. In an ideal fluid, the hydrodynamic damping matrix, $D(\dot{q})$, is real, nonsymmetrical and strictly positive. With rough assumptions such as a symmetric robot and non-coupled motion, it can be simplified to a diagonal matrix $diag(d1 + d2|\dot{q})|_i\ i = 1,...,6$ where $\boldsymbol{d_1}$ is a linear damping coefficient and $\boldsymbol{d_2}$ is a quadratic (drag) damping coefficient. In the hydrodynamic terminology, the gravitational and buoyant forces are called restoring forces, **G(*x*)**. The gravitational forces will act through the center of gravity while the buoyant forces act through the center of buoyancy. Environmental disturbances, ***w***, due to waves, wind, and ocean currents and their mathematical expressions are discussed in detail in Fossen (1994). Components of the control matrix, ***B***, are dependent on each robot's configuration, control surfaces, number of thrusters, and thruster locations. Therefore, ***B*** may not be a square matrix. The thruster force, ***u***, will be the output of each thruster whose dynamics are nonlinear and quite complex.
Difficulty to find several parameters on modeling the dynamics motion of vehicle likes Hull resistance, Moment, Lift at each different angel of attack, etc. could be helped with Precise Computational Fluid Dynamics (CFD) Software. This's the results of CFD Analysis to AUV-Sotong to find moment, Coefficient Moment, distribution dynamics pressure, and velocity (show on **figure 10**).

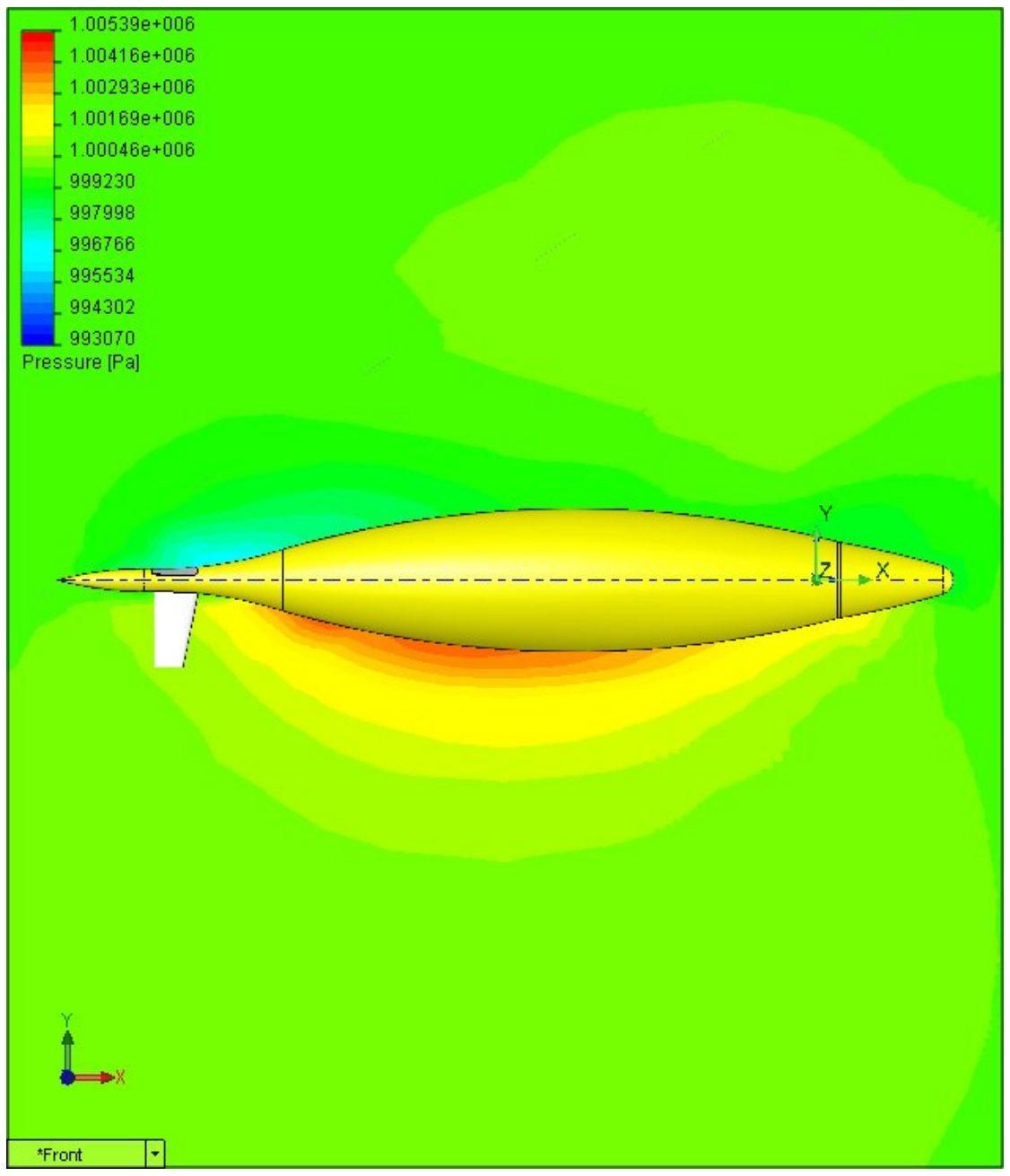

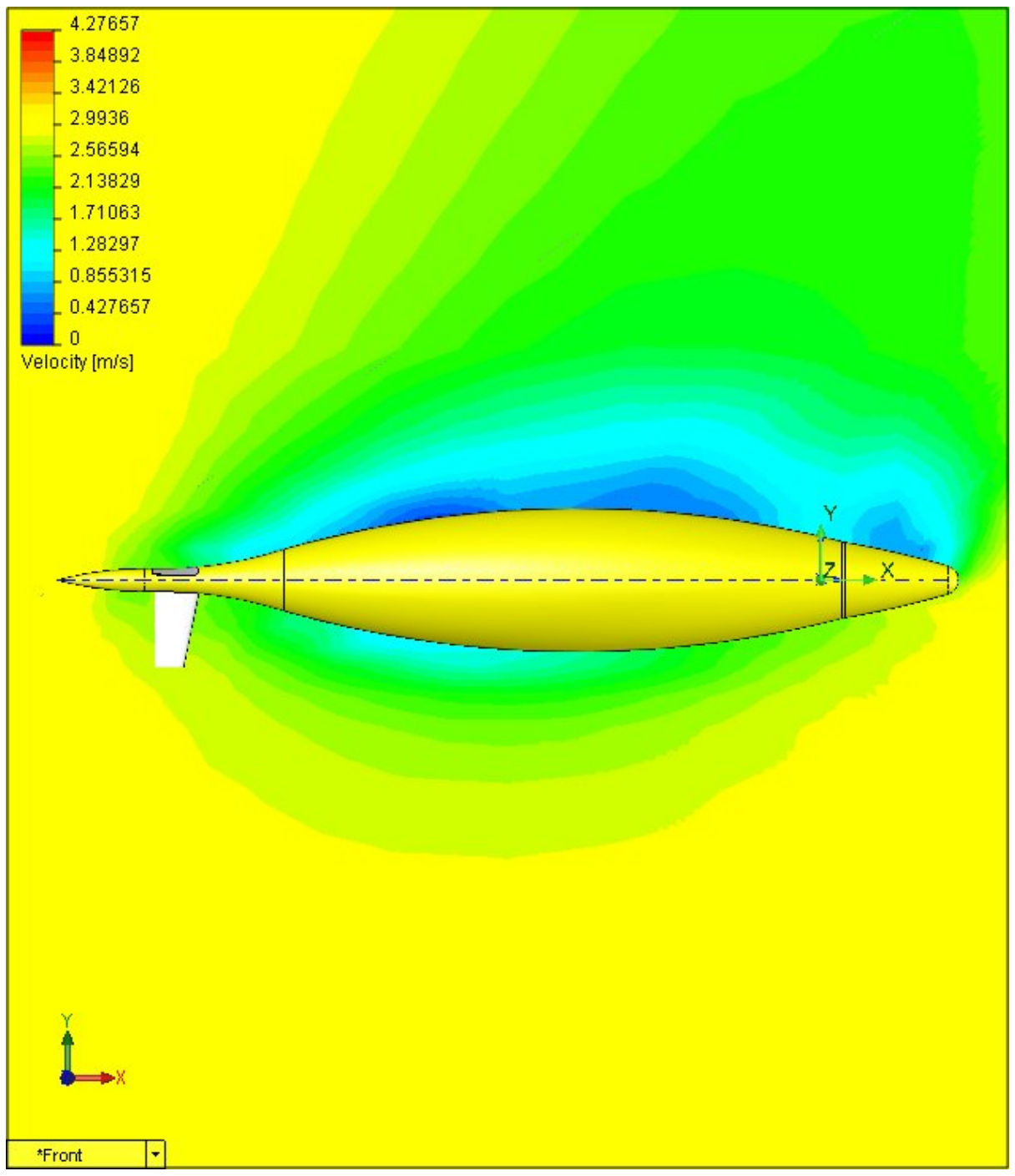

| Name | Unit | Value | Pro gre ss | Use in conver gence | Delta | Criteri a |
|---|---|---|---|---|---|---|
| Momen t | N*m | -7814.38 | 100 | On | 60.573 3749 | 61.888 5453 |
| Cm | N*m | -0.224474 | 100 | On | 0.0017 40020 4 | 0.0017 77799 76 |

Figure 10. CFD Analysis AUV Sotong Hull at -2 deg Angle of attack

## 5 Conclusions

The paper present the progress of research and development of autonomous Underwater Vehicle in this case on Bandung institute of Technology. Some of progress especially on field testing being attention.
The technical specification and control architecture of the prototypes are presented as illustration above. This testing focused on heading lock maneuver with Classical PID Controller. This methods reports to be successfully applied wit this system. The advanced development of controller such as sliding control (Yoerger and Slotine, 1985; Healey and Lienard, 1993), nonlinear control (Nakamura and Savant, 1992), adaptive control (Goheen et al., 1990; Yuh, 1990a, 1996; Cristi et al., 1991; Tabaii et al., 1994, Choi and Yuh, 1996; Nie et al., 1998), neural network control (Yuh, 1990b, 1994; Lorenz andYuh, 1996; Ishii et al., 1998), and fuzzy control (DeBitetto, 1994; Kato, 1995) will be apply for the next generations.

And consequentially of this development, the field testing of AUV will be more improved. The reliability and efficiency of the operation AUV-Sotong will be centered.

## 6 Acknowledgment

The work was supported by the Research Grant made available by the Technology Assessment and Application Agency (BPPT). The authors would like to thank to the technical team involved in through the design process, manufacturing and testing of the vehicle.